\documentclass[runningheads]{llncs}
\usepackage{graphicx}
\usepackage{amsmath,amssymb} 
\usepackage{color}
\usepackage[width=122mm,left=12mm,paperwidth=146mm,height=193mm,top=12mm,paperheight=217mm]{geometry}

\usepackage{subcaption}
\usepackage{multirow}
\usepackage{floatrow}  
\newfloatcommand{capbtabbox}{table}[][\FBwidth]

\begin{document}
\pagestyle{headings}
\mainmatter

\title{Learning Common and Specific Features for RGB-D Semantic Segmentation with Deconvolutional Networks} 

\titlerunning{Learn Common and Specific Features for RGB-D Semantic Segmentation}

\authorrunning{J. Wang, Z. Wang, D. Tao, S. See, and G. Wang}

\author{Jinghua Wang$^\dag$, Zhenhua Wang$^\dag$, Dacheng Tao$^\ddag$, Simon See$^\S$, Gang Wang$^{\dag*}$}


\institute{$^\dag$Nanyang Technological University $^*$Corresponding Author \\
	$ ^\ddag $ University of Technology Sydney (UTS) $ ^\S $ NVIDIA Corporation
	\email{jinghuawng@gmail.com,zhwang.me@gmail.com, dacheng.tao@uts.edu.au,ssee@nvidia.com,wanggang@ntu.edu.sg$^*$}
}

\maketitle

\begin{abstract}
In this paper, we tackle the problem of RGB-D semantic segmentation of indoor images. We take advantage of deconvolutional networks which can predict pixel-wise class labels, and develop a new structure for deconvolution of multiple modalities. We propose a novel feature transformation network to bridge the convolutional networks and deconvolutional networks. In the feature transformation network, we correlate the two modalities by discovering common features between them, as well as characterize each modality by discovering modality specific features. 
With the common features, we not only closely correlate the two modalities, but also allow them to borrow features from each other to enhance the representation of shared information. 
With specific features, we capture the visual patterns that are only visible in one modality.
The proposed network achieves competitive segmentation accuracy on NYU depth dataset V1 and V2.

\end{abstract}

\section{Introduction}

\begin{figure}[t]
	\begin{center}
		\includegraphics[width=0.8\linewidth]{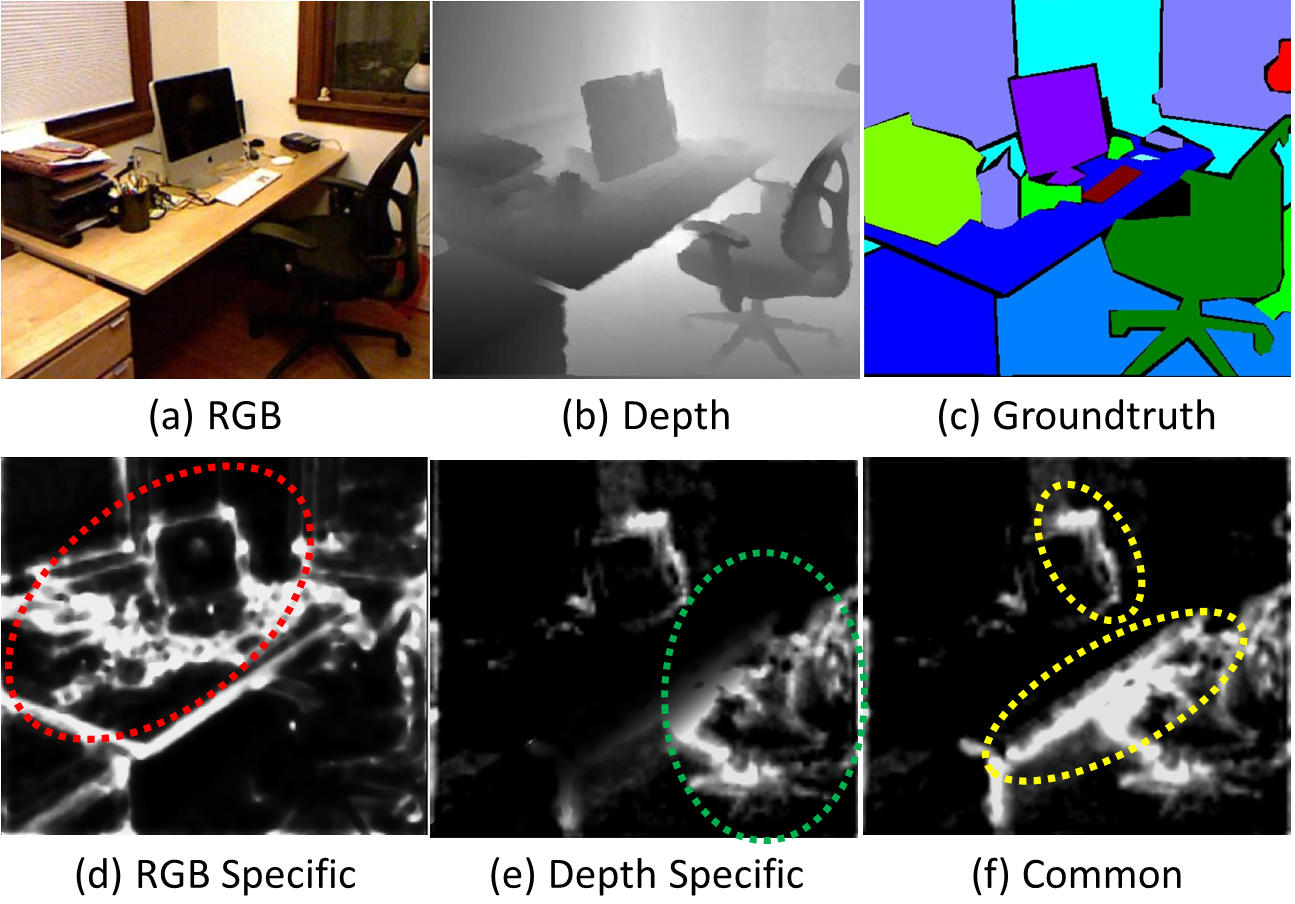}
	\end{center}
	\caption{Example images from the NYU Depth Dataset V2 \cite{Silberman:ECCV2012:Indoor_Segmentation}. (a) shows an RGB image captured in a homeoffice. (b) and (c) are the corresponding depth map and groundtruth. (d-f) are the visualized RGB specific feature, depth specific feature, and common feature (The method to obtain these features will be discussed in Section \ref{sec:experiment-testing}.). 
		RGB specific features encode the texture-rich visual patterns, such as the objects on the desk (the red circle in (d)). The depth specific features encode the visual patterns which are more obvious in the depth map, such as the chair (the green circle in (e)). Common features encode the visual patterns that are visible in both modalities, such as the edges (the yellow circles in (f))}
	\label{fig:difficult_examples}
\end{figure}

Semantic segmentation of scenes is a fundamental task in image understanding. It assigns a class label to each pixel of an image. Previously, most research works focus on outdoor scenarios \cite{ParsingNaturalScenes_SocherEtAl2011:RNN},\cite{Shuai2015DAG}, \cite{Farabet_2013LearningHierarchicalFeatures_TPAMI_}, \cite{Hong_NIPS2015_Decoupled}, \cite{Shuai2016Scene},\cite{noh_ICCV2015_learning}. 
Recently, the semantic segmentation of indoor images attracts increasing attention  \cite{Silberman:ECCV2012:Indoor_Segmentation}, \cite{Silberman_ICCV2011_Indoor_scene},
\cite{Farabet_2013LearningHierarchicalFeatures_TPAMI_},   \cite{Ren_Xiaofeng_CVPR2012_RGB-D_SceneLabeling}, \cite{Gupta_CVPR2013_Perceptual_Organization}, \cite{couprie_2013ICLR_indoor_semantic}, \cite{Khan_ECCV2014_Geometry}, \cite{gupta_ECCV14_Learning_rich}, \cite{Deng_ICCV2015_Semantic_segmentation}, \cite{Banica2015Second}.
It is challenging due to many reasons, including randomness of object distribution, poor illumination, occlusion and so on. 
Fig. \ref{fig:difficult_examples} shows an example of indoor scene segmentation.

Thanks to the Kinect and other low-cost RGB-D cameras, 
we can obtain not only the color images (Fig. \ref{fig:difficult_examples} (a)), but also the depth maps of indoor scenes (Fig. \ref{fig:difficult_examples} (b)). The additional depth information is independent of illumination, which can significantly alleviate the challenges in semantic segmentation.
With the availability of RGB-D indoor scene datasets \cite{Silberman:ECCV2012:Indoor_Segmentation}, \cite{Silberman_ICCV2011_Indoor_scene}, many methods \cite{Farabet_2013LearningHierarchicalFeatures_TPAMI_},   \cite{Ren_Xiaofeng_CVPR2012_RGB-D_SceneLabeling}, \cite{Gupta_CVPR2013_Perceptual_Organization}, \cite{Wang2015Unsupervised}, \cite{couprie_2013ICLR_indoor_semantic}, \cite{Khan_ECCV2014_Geometry}, \cite{gupta_ECCV14_Learning_rich}, \cite{Deng_ICCV2015_Semantic_segmentation} are proposed to tackle this problem. These methods can be divided into two categories according to how they learn appropriate features to represent the visual patterns.
While the methods \cite{Silberman:ECCV2012:Indoor_Segmentation}, \cite{Silberman_ICCV2011_Indoor_scene}, \cite{Ren_Xiaofeng_CVPR2012_RGB-D_SceneLabeling},  \cite{Gupta_CVPR2013_Perceptual_Organization},  \cite{Deng_ICCV2015_Semantic_segmentation} rely on  low level or hand-crafted features to produce the label map, the works \cite{Farabet_2013LearningHierarchicalFeatures_TPAMI_}, \cite{Shuai2015Integrating}, \cite{couprie_2013ICLR_indoor_semantic}, \cite{gupta_ECCV14_Learning_rich}, \cite{Wang_ECCV2014_multi-modal-unsupervised}, \cite{Wang2015MMSS}, \cite{Shuai2015Quaddirectional} learn deep features based on CNN (convolutional neural networks).

To apply CNN-based method on two modalities (RGB and depth) semantic segmentation, we can train two independent CNN models for RGB images and depth maps, then simply combine them together by decision score fusion. However, this strategy ignores the correlation between these two modalities in feature learning. To capture the correlation between different modalities, the previous methods
\cite{Farabet_2013LearningHierarchicalFeatures_TPAMI_}, \cite{couprie_2013ICLR_indoor_semantic}, \cite{gupta_ECCV14_Learning_rich}, \cite{Wang_ECCV2014_multi-modal-unsupervised} concatenate the RGB image with the depth map to form a four-channel signal and take them as the input.
As pointed out in \cite{ngiam_2011ICML_multimodal}, these methods can only capture the shallow correlations between two modalities. In the learned network structure, most of the hidden units only have strong connections with a single modality. In addition, the modality specific features, which are very useful to characterize one particular modality, are heavily suppressed. For example, to segment the objects on the desk in Fig. \ref{fig:difficult_examples}, we can learn discriminative features only from the RGB image. If we concatenate the RGB image and depth map, we are more likely to learn the common features that are visible in both modalities, and lose the RGB specific features to encode textures.

To learn informative features from both RGB image and depth map, we propose to correlate these two modalities by discovering 
their common features while characterize each modality by exploiting its specific features.
To achieve this, we introduce a new network structure as an extension of deconvolutional network \cite{noh_ICCV2015_learning} for RGB-D semantic segmentation.
Fig. \ref{fig:CNNStructure} shows the overall structure of the proposed model.
The model has a convolutional network and deconvolutional network for each modality, as well as a novel feature transformation network to bridge them.
Specifically, the convolutional networks extract features for each modality.
The feature transformation network disentangles common features and modality-specific features from the top-layer covolutional features of each modality. The common features (Fig. \ref{fig:difficult_examples} (f)), which represent deep correlations between two modalities, are expected to encode information shared by both modalities. The specific features (Fig. \ref{fig:difficult_examples} (d) and (e)) are expected to encode information that is visible in only one modality.
A separate deconvolutional network is used to predict the decision score for each modality, which receives the common and specific features of its corresponding modality and the common features borrowed from the other modality. 
Finally, the label map is obtained by decision score fusion.

It is worth noting that we explicitly allow one modality to borrow common features learned from other modality to enhance the representation of their shared information.
Such a compensation is quite useful especially when the data from one modality is not well captured.

The contribution of this work is mainly twofold. Firstly, we introduce deconvolutional neural network for multimodal semantic segmentation.
Secondly, we develop a framework to model common and specific features
to enhance the segmentation accuracy. With the learned common feature, the two modalities can help each other to generate robust deconvolutional features.

The rest of this paper is organized as follows. Section \ref{sec:relatedWork} reviews the related work. Section \ref{sec:approach} presents our network architecture.
Section \ref{sec:training} presents our training method.
Section \ref{sec:experiment} shows our experiments. Section \ref{sec:conclusion} concludes this paper.

\section{Related Work}
\label{sec:relatedWork}

Multi-modality feature learning is widely studied these days.
Socher et al. \cite{ParsingNaturalScenes_SocherEtAl2011:RNN} introduce recursive neural networks (RNNs)
for predicting recursive structure in two different modalities, i.e. the image and the natural language. The proposed RNNs model can not only identify the items inside an image or a sentence but also capture how they interact with each other.
Farabet et al. \cite{Farabet_2013LearningHierarchicalFeatures_TPAMI_} introduce multi-scale convolutional neural networks to learn dense feature extractors. The proposed multi-scale representations successfully capture shape and texture information, as well as the contextual information. However, this method cannot generate cleanly delineated predictions without post-processing.
Ngiam et al. \cite{ngiam_2011ICML_multimodal} propose bimodal deep auto-encoder to learn more representative shared features from multiple modalities. This work also demonstrates that we can improve the feature learning of one modality if multiple modalities are available at the training time.
By introducing a domain classifier, Ganin and Lempitsky \cite{ganin15_icml2015_unsupervised} learn domain invariant features based on labeled data from source domain and unlabeled data from target domain.
In order to generate one modality from the other, Sohn et al. \cite{sohn-NIPS2014-improved-multimodal} propose to use information variation as the objective function in a multi-modal representation learning framework. To learn transferable features in high layers of the neural network, Long et al. \cite{Long_icml2015_learning-transfer} propose a deep adaption network to minimize the maximum mean discrepancy of the features.

Thanks to the low-cost RGB-D camera, we can obtain not only RGB but also depth information to tackle semantic segmentation of indoor images.
Koppula et al. \cite{Koppula_NIPS2011_SemanticLabeling} use graphical model to capture contextual relations of different features. This method is computationally expensive as it relies on the 3D+RGB point clouds.
Ren et al. \cite{Ren_Xiaofeng_CVPR2012_RGB-D_SceneLabeling} propose to first model appearance (RGB) and shape (depth) similarities using kernel descriptors, then capture the context using superpixel Markov random field (MRF) and segmentation tree.
Couprie et al. \cite{couprie_2013ICLR_indoor_semantic} extend the multi-scale convolutional neural network \cite{Farabet_2013LearningHierarchicalFeatures_TPAMI_} to learn multi-modality features for semantic segmentation of indoor scene. 
Wang et al. \cite{Wang_ECCV2014_multi-modal-unsupervised} propose an unsupervised learning framework that can jointly learn visual patterns from RGB and depth information.
Deng et al. \cite{Deng_ICCV2015_Semantic_segmentation} introduce mutex constraints in conditional random field (CRF) formulation to eliminate the configurations that violate common sense physics laws.

Long et al. \cite{long_shelhamer_CVPR2015_fcn}  propose fully convolutional networks (FCN) that can produce a label map which has the same size of the input image. 
FCN is an extension of CNN  \cite{Krizhevsky_NIPS2012_ImageNet_classification} by interpreting the fully connected layers as convolutional layers with large receptive fields.  As FCN can be trained end-to-end, and pixels-to-pixels, it can be directly used for the task of semantic segmentation. However, FCN has two disadvantages: 1) it cannot handle various scales of semantics; 2) it loses many detailed structure of the object. 
To overcome these limitations, Noh et al. \cite{noh_ICCV2015_learning} propose to train deconvolutional neural networks based on VGG net for semantic segmentation. Papandreou et al. \cite{papandreou_arxiv_2015_weakly} propose a method to learn deconvolutional neural networks from weakly annotated training data. Hong et al. \cite{Hong_NIPS2015_Decoupled} decouple the tasks of classification and segmentation by modeling them with two different networks and a bridging layer to connect them.
Deconvolutional networks can be considered as the reverse process of the convolutional network. It explicitly reconstructs the label map through a series of deconvolutional and unpooling layers.
It is suitable for generating dense and precise label maps.
Compared with the other CNN-based methods of semantic segmentation, deconvolutional networks \cite{noh_ICCV2015_learning} are more efficient as they can directly produce the label map.

\section{Approach}
\label{sec:approach}

In the task of RGB-D indoor semantic segmentation, the inputs are the RGB image and the corresponding depth map. The output is the semantic label map, i.e. the class label of every pixel.

Instead of conducting segmentation based on the pixel values, we learn informative representations from regions of these two modalities.
The benefit of using multiple modalities is not limited to the fact that one modality can cover the shortage of the other.  As stated by Ngiam et al. \cite{ngiam_2011ICML_multimodal}, we can improve the feature extraction procedure of one modality with the help from data of another modality. 
On one hand, as some visual patterns are visible in both modalities, we expect to extract a set of similar features from the RGB image and the corresponding depth map. 
On the other hand, as the RGB image mainly captures the appearance information and the depth map mainly captures shape information, we expect to extract some modality-specific features for each of them.

In this work, we explicitly learn common features and modality-specific features for both modalities.
By jointly maximizing similarities between shared information and differences between modality-specific information, we learn to disentangle features of each modality into common features and specific features respectively.
To achieve robust prediction, we explicitly allow one modality to borrow common features learned from other modality to enhance the representation of their shared information. Such a mechanism is quite useful especially when the data from one modality is not well captured.
The final result is obtained by fusing decision scores of the two modalities.

\begin{figure}[t]
	\begin{center}
		\includegraphics[width=0.9\textwidth]{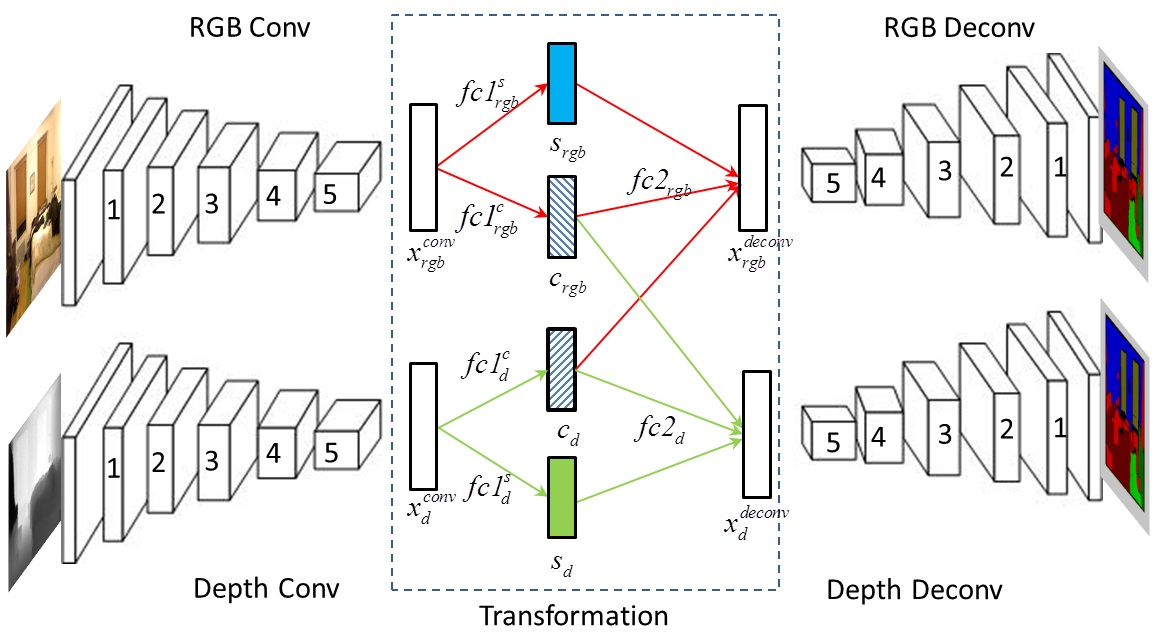}
	\end{center}
	\caption{Overall structure of the proposed network. The RGB and depth convolutional network have the same structure, consisting of $ 14  $ convolutional layers and $ 5 $ pooling layers. The deconvolutional networks are the mirrored version of the convolutional networks. The last layer of the convolutional network (i.e. conv 6 in Tab. \ref{tab:layer-size}) produce the convolutional features $ x_{rgb}^{conv} $ and $ x_{d}^{conv} $. Based on  $ x_{rgb}^{conv} $ and $ x_{d}^{conv} $, our feature transformation network learns to extract common features $ c_{rgb} $ (or $ c_d $) by fully connected layer $ fc1_{rgb}^c $ (or $ fc1_{d}^c $), and modality specific features $ s_{rgb} $ (or $ s_d $) by fully connected layer $ fc1_{rgb}^s $ (or $ fc1_d^s $). To obtain robust deconvolutional features, the fully connected layer $ fc2_{rgb} $  takes three types of feature as input: RGB-based features ($ c_{rgb}  $ and $ s_{rgb} $), as well as the borrowed common feature ($ c_d $) from depth modality. Similarly, the layer $ fc2_{d} $ also takes three features as input}
	\label{fig:CNNStructure}
\end{figure}

\subsection{Network Structure}

As shown in Fig \ref{fig:CNNStructure}, our network has five components: RGB convolutional network, depth convolutional network, feature transformation network, RGB deconvolutional network, and depth deconvolutional network.
Both of two convolutional networks are designed based on the VGG16 net [22]. Specifically, each convolutional network has 14 convolutional layers (with corresponding ReLU and pooling layers between them). The two deconvolutional networks are mirrored versions of the convolution networks, each of which has multiple unpooling, deconvolutional and ReLU layers. The feature transformation network
lies in-between the convolutional and deconvolutional networks, which consists of several fully connected layers. 
Tab. 1 shows the detailed configurations of our network. Note that we only show the networks for RGB modality.


\begin{table}[!t]
	\caption{Detailed configuration of the network. (a) shows the RGB convolutional network. (b) shows the deconvolutional network.  (c) shows the feature transformation network.  We use \textit{conv (deconv)} to denote convolutional (deconvolutional) layers, and \textit{pool (unpool)} to denote pooling (unpooling) layers. The layer conv 6 produces the convolutional features. The fully connected layers  $ fc1_{rgb}^{s} $, $ fc1_{rgb}^{c} $,  $ fc2_{rgb} $ respectively produces the RGB modality specific features, RGB common features, and RGB deconvolutional features}
	\label{tab:layer-size}
	\begin{subtable}{.5\linewidth}
		\centering
		\caption{Convolutional network}
		\begin{tabular}{ | c | c | c |}
			\hline
			name &  kernel  & output size\\ \hline
			\hline
			image & - & $480 \times 640$ \\ \hline
			conv 1: 1-2 & $ 3 \times 3 $  & $480 \times 640 \times 64$ \\ 
			pool 1  & $ 2 \times 2 $  & $240 \times 320 \times 64$ \\ \hline
			conv 2: 1-2 & $ 3 \times 3 $  & $240 \times 320 \times 128$ \\ 
			pool 2  & $ 2 \times 2 $  & $120 \times 160 \times 128$ \\ \hline
			conv 3: 1-3 & $ 3 \times 3 $  & $120 \times 160 \times 256$ \\ 
			
			pool 3  & $ 2 \times 2 $  & $60 \times 80 \times 256$ \\ \hline
			conv 4: 1-3 & $ 3 \times 3 $  & $60 \times 80 \times 512$ \\ 
			
			pool 4  & $ 2 \times 2 $  & $30 \times 40 \times 512$ \\ \hline
			conv 5: 1-3 & $ 3 \times 3 $  & $30 \times 40 \times 512$ \\ 
			pool 5  & $ 2 \times 2 $  & $ 15 \times 20 \times 512$ \\
			
			conv 6  & $ 15 \times 20 $  & $ 1 \times 1 \times 4096$ \\ 
			\hline 
	
		\end{tabular}  
		
		
		\begin{tabular}{ | c | c | c |}
			\multicolumn{3}{c}{(c) Transformation network} \\
			\hline
			name &  kernel  & output size\\ \hline
			\hline
			$ fc1_{rgb}^{s} $ & $ 1 \times 1 $ & $ 1 \times 1 \times 4096$ \\ 
			$ fc1_{rgb}^{c} $ & $ 1 \times 1 $ & $ 1 \times 1 \times 4096$ \\ 
			$ fc2_{rgb} $ & $ 1 \times 1 $ & $ 1 \times 1 \times 4096$ \\ 
			\hline 
		\end{tabular}

	\end{subtable}%
	\begin{subtable}{.5\linewidth}
		\centering
		\caption{Deconvolutional network}
		\begin{tabular}{ | c | c | c |}

			\hline
			name &  kernel & output size\\ \hline
			\hline
			deconv 6  & $ 15 \times 20 $  & $ 15 \times 20 \times 4096$ \\ 
			unpool 5  & $ 2 \times 2 $  & $ 30 \times 40 \times 512$ \\ 
			deconv 5: 1-3 & $ 3 \times 3 $  & $30 \times 40 \times 512$ \\ \hline
			unpool 4  & $ 2 \times 2 $  & $60 \times 80 \times 512$ \\ 
			deconv 4: 1-2  & $ 3 \times 3 $  & $60 \times 80 \times 512$ \\ 
			deconv 4: 3 & $ 3 \times 3 $  & $60 \times 80 \times 256$ \\
			\hline
			unpool 3  & $ 2 \times 2 $  & $120 \times 160 \times 256$ \\ 
			deconv 3: 1-2 & $ 3 \times 3 $  & $120 \times 160 \times 256$ \\ 
			deconv 3:  3 & $ 3 \times 3 $  & $120 \times 160 \times 512$ \\ \hline
			
			unpool 2  & $ 2 \times 2 $  & $240 \times 320 \times 128$ \\ 
			deconv 2: 1 & $ 3 \times 3 $  & $240 \times 320 \times 128$ \\
			deconv 2:  2 & $ 3 \times 3 $  & $240 \times 320 \times 64$ \\ \hline
			unpool 1  & $ 2 \times 2 $  & $480 \times 640 \times 64$ \\ 
			deconv 1: 1-2 & $ 3 \times 3 $  & $480 \times 640 \times 64$ \\ \hline
			\hline
			label map & $ 1 \times 1 $  & $480 \times 640 \times 14$ \\ \hline      
		\end{tabular}  
	\end{subtable} 
\end{table}

Connecting the convolutional and deconvolutional networks  is the feature transformation network, which takes the convolutional features as input and produces the deconvolutional features as output.
In Tab. \ref{tab:layer-size}, the convolutional layer  conv 6  generates the RGB convolutional features $ x_{rgb}^{conv} $, which are transformed into common feature $ c_{rgb}$  by fully connected layers $ fc1_{rgb}^c $   and modality specific feature $ s_{rgb} $  by layer $ fc1_{rgb}^s $.

We expect the common features from two different modalities to be similar to each other while the specific features to be different to each other. Hence, we propose to use multiple kernel maximum mean discrepancy (which will be discussed later in sec. \ref{sec:mkmmd}) to access these similarities and differences. 
To obtain robust deconvolutional features, we allow one modality to borrow the common features from the other. As shown in Fig \ref{fig:CNNStructure}, the fully connected layer $ fc2_{rgb} $ produces the RGB deconvolutional features by taking the RGB modality specific feature $ s_{rgb} $ and both of the common features ($ c_{rgb} $ and $ c_d $) as inputs. Similarly, the layer $ fc2_{d} $ transform $ s_d $, $ c_d $, and $ c_{rgb} $  into depth deconvolutional features.

In this framework, the two modalities can boost each other with the learned common features. 
It is helpful when the data of one modality is poorly captured and loses some key information.
As the data from different modalities is captured using different mechanisms, one modality is expected to provide complementary information to the other.

The RGB (depth) deconvolutional network is the mirrored version of the RGB (depth) convolutional network.  Each convolutional (pooling) layer in convolutional network has a corresponding deconvolutional (unpooling) layer. The unpooling layers of the RGB (depth) deconvolutional network use the pooling masks learned in RGB (depth) convolutional network.
While the pooling layers gradually reduce the size of the feature map, the unpooling layers gradually enlarge the feature maps to obtain precise label map.

Unpooling can be considered as a reverse process of pooling.
Pooling is a strategy of sub-sampling by selecting the most responsive node in the region of interest. 
Mathematically, pooling is an irreversible procedure. However, we can record the location of the most responsive node by a mask and use this mask to recover the activation to its right place in the unpooling layer. Note that the  RGB and depth deconvolutional network use different pooling masks learned by their corresponding convolutional networks.
The unpooling layer can produce a sparse feature map representing the main structure.

Taking a single activation as input, the filters in a deconvolutional layer produce multiple outputs.
Based on the sparse un-pooled feature map, deconvolutional layers reconstruct the details of the label map through convolution-like operations but in a reverse manner.
A series of deconvolutional layers hierarchically capture different level of the shape information.  
Higher layer corresponds to more detailed shape structure.

\subsection{Multiple kernel maximum mean discrepancy (MK-MMD)}
\label{sec:mkmmd}

This section introduces the measurement to assess the similarity between common features and modality specific features.
To obtain similar RGB and depth common feature, we may simply minimize their Euclidean distance.
However, Euclidean distance is sensitive to outliers which don't share very similar 
common features.
We can overcome this limitation by considering the common (specific) features of two modalities as samples from two distributions and calculating the distance between the distributions. 
We aim to obtain two similar distributions for common features and different distributions for specific features. 
If most of RGB common features and depth common features are similar, we may conclude that their distributions are similar, even if they are significantly different for a few noisy outliers.

Hence, we do not expect the common features $ c_d $ and $ c_{rgb} $ (output of the layer $ fc1_d^c $ and $ fc1_{rgb}^c $ in Fig \ref{fig:CNNStructure}) of two different modalities to be the same individually. Instead, we adopt the MK-MMD to assess the similarity between their distributions.

Given a set of independent observations from two distributions $ p $ and $ q $, two-sample testing accepts or rejects the null hypothesis $ H_0: p=q $, i.e. the distributions that generate these two sets of observations are the same. The acceptance or rejection is made based a certain test statistic.

There are many existing techniques to calculate the similarity between distributions, 
such as entropy, mutual information, or KL divergence. However, these information theoretic approaches rely on the density estimation, or sophisticated space-partitioning/bias-correction strategies which are typically infeasible for high-dimensional data.

The kernel embedding allows us to represent a probability distribution as an element of a reproducing kernel Hilbert space.
Let the kernel function $ k $ define a reproducing kernel Hilbert space $ F_k $ in a topological space $ X $. The mean embedding of distribution $ p $ in Hilbert space $ F_k $ is a unique element $ \mu_k(p) $ such that \cite{Berlinet_2004_Reproducing}:
\begin{equation}
E_{x \sim p}f(x)=<f(x),\mu_k(p)>_{H_\phi}, \qquad \forall f \in {H_k}.
\end{equation}
As stated by Riesz representation theorem, the mean embedding $ \mu_k $ exists if the kernel function $ k $ is Borel-measurable and $ E_{x \sim p}k^{1/2}(x,x)<\infty $.

As a popular test statistic in two-sample testing, MMD (maximum mean discrepancy) calculates the norm of the difference between embeddings of two different distributions $ p $ and $ q $, as follows
\begin{equation}
MMD(p,q)=\| \mu_k(p) - \mu_k(q)\|^2_{F_k}.
\end{equation}
In theory, MMD equals to the upper bound of the difference in expectations between two probability distributions, i.e.
\begin{equation}
MMD(p,q)= \sup_{\|f\|_H \leq 1} \| E_p[f(p)] - E_q[f(q)]\|.
\end{equation}

MMD is heavily dependent on the choice of kernel function $ k $. In other words, we may obtain contradictory results using two different kernel functions. 
Gretton et al. \cite{Gretton_NIPS2012_Optimal_kernel} propose MK-MMD (multiple kernel maximum mean discrepancy) in two-sample testing, which can minimize the Type II error (false accept $ p=q $) given an upper bound on Type I error (false reject $ p=q $ ).
By generating a kernel function based on a family of kernels, MK-MMD can improve the test power and is successfully applied to domain adaption \cite{Long_icml2015_learning-transfer}.
The kernel function $ k $ in MK-MMD is a linear combination of positive definite functions $ \{k_u\} $, i.e.
\begin{equation}
\Bbbk:=\{k=\sum_{u=1}^{d} \beta_u k_u|\sum_{u=1}^{d} \beta_u=D>0; \beta_u \geq 0, \forall u \}.
\label{eq:linear-combine-of-multiple-kernels}
\end{equation}

The distance between two distributions calculated based on MK-MMD can be formulated as follows
\begin{equation}
d(p,q)=\| \mu_k(p) - \mu_k(q)\|^2_{F_k}= \sum_{u=1}^{d}{\beta_u d_u(p,q)}.
\label{eq:mkmmd-formulation}
\end{equation}
where $ d_u(p,q) $ is the MMD for the kernel function $ k_u $.

In the training stage, we use the following function to calculate the unbiased estimation of MK-MMD between the common features
\begin{equation}
\begin{aligned}
d(c_{rgb},c_d)&=\frac{2}{n} \sum_{i=1}^{\left. n \middle / 2 \right.}{\eta (u_i)}.\\
\eta (u_i)= & k(c_{rgb}^{2i-1},c_{rgb}^{2i})-k(c_{rgb}^{2i-1},c_d^{2i})\\
+ & k(c_d^{2i-1},c_d^{2i})-k(c_d^{2i-1},c_{rgb}^{2i}).\\
\end{aligned}
\label{eq:MK-MMD-common_specific}
\end{equation}
where $ n $ is the batch size, $ c_{rgb}^{i} $ and $ c_{d}^{i} $ ($ 1 \leq i \leq n $) are the RGB common feature and depth common feature respectively produced by layer $ fc1_{rgb}^c $ and $ fc1_{d}^c $.
Also, similar to Eq. \ref{eq:MK-MMD-common_specific}, we can calculate the similarity between the RGB modality specific feature $ s_{rgb} $ (produced by $ fc1_{rgb}^s $) and depth specific feature $ s_{d} $ (produced by $ fc1_{d}^s $).


In our framework, the common features $ c_{rgb} $ and $ c_d $ are expected to be similar to each other as much as possible. The modality specific features $ s_{rgb} $ and $ s_d $ are expected to different from each other. Thus, we try to minimize $ d(c_{rgb},c_d) $ and simultaneously maximize $ d(s_{rgb},s_d) $.
The loss function of our network is as follows
\begin{equation}
\label{eq:lossfunction}
L=\alpha_{rgb} l_{rgb}+\alpha_{d}l_d+\alpha_{c}d(c_{rgb},c_d)-\alpha_{s}d(s_{rgb},s_d).
\end{equation}
where $ l_{rgb} $ and $ l_d $ are the pixel-wise losses between the label map and the outputs of the deconvolutional network. We use the parameters $ \alpha_{rgb}, \alpha_{d}, \alpha_{c} $, and $ \alpha_s $ to balance the four terms.
In the back propagation, the gradient of the common and modality specific features are calculated from two different sources: the deconvolutional features and the MK-MMD distances.

\section{Training}
\label{sec:training}

Following the work \cite{noh_ICCV2015_learning}, we adopt a two-stage method to train our network.  
In the first stage, we train our network using image patches containing a single object, and learn how to segment an object from its surroundings. 
In the second stage, we generate patches based on bounding box proposals \cite{ZitnickDollar_ECCV14_edgeBoxes}. The generated patches contain two or more objects. In this stage, we train the network to learn how to segment two or more neighboring objects.

The kernel function in Eq. \ref{eq:MK-MMD-common_specific} is a linear combination of $ d $ different Gaussian kernels (i.e. $ k_u(x,y)= e^{-\|x-y \|^2/\sigma_u}$). In our experiment, we use $ 11 $ kernel functions, i.e. $ d=11 $ in Eq. \ref{eq:MK-MMD-common_specific}. The $ \sigma_u $ is set to be $ 2^{u-6} (u=1,\dots, 11) $ .
We observe that $ 11 $ kernel functions are sufficient to disentangle common features and specific features in our task.
The parameter $\beta$ in Eq. \ref{eq:mkmmd-formulation} is learned based on the method proposed in Gretton \cite{Gretton_NIPS2012_Optimal_kernel}, and the values are [2,3,9,12,14,15,15,14,10,5,1]*1e-2. We learn the four parameters in Eq. \ref{eq:lossfunction} by cross-validation.

We implement our network based on caffe \cite{jia2014caffe} and the deconvolutional network \cite{noh_ICCV2015_learning}. We employ the standard stochastic gradient descent with momentum for optimization. In the training stage, while the convolutional networks are initialized using the VGG 16-layer net \cite{Simonyan14c_VGG_verydeep} pre-trained on ILSVRC dataset \cite{Deng09imagenet:a}, the deconvolutional networks are initialized randomly. We set the learning rate, weight decay and momentum respectively to be $ 0.01 $, $ 0.0005 $ and $ 0.9 $.

In this work, we decompose the deconvolutional network into five components based on the size of feature map and train one component after another. Following \cite{eigen2015predicting}, we train the network by predicting a coarse output for each component.  
For example, we train the first component (from deconv 6 to deconv 5-3) to predict the downsampled ($ 30 $ by $ 40 $) label map.

\section{Experiments}
\label{sec:experiment}

Two popular RGB-D datasets for semantic segmentation of indoor scene images are NYU Depth dataset V1 \cite{Silberman_ICCV2011_Indoor_scene} and NYU Depth dataset V2 \cite{Silberman:ECCV2012:Indoor_Segmentation}. The $ 2,347 $ RGB-D images in dataset V1 are captured in $ 64 $ different indoor scenes. As in the work \cite{Silberman_ICCV2011_Indoor_scene}, we group the $ 1,518 $ different names into $ 13 $ categories, i.e.\textit{ bed, blind, book, cabinet, ceiling, floor, picture, sofa, table, tv, wall, window,others}. 
The $ 1,449 $ RGB-D images in dataset V2 are captured in $ 464 $ different indoor scenes.
Following \cite{Silberman:ECCV2012:Indoor_Segmentation}, we group the $ 894 $ different names into $ 13 $ categories, i.e. \textit{bed, objects, chair, furniture, ceiling, floor, decorate, sofa, table, wall, window, books, and TV}.

\subsection{Baselines}
\label{sec:experiment-baseline}

To show the effectiveness of the proposed method, we compare it with five baselines. In the first baseline (B-DN), we train two deconvolutional networks independently, each takes the convolutional features of one modality as the input. The final segmentation results are obtained by decision score fusion.
In the second baseline (S-DN), we have two convolutional networks and one deconvolutional network. The deconvolutional features are transformed directly from the concatenation of convolutional features from two  modalities.  
The previous two baselines do not consider the correlations between these two modalities explicitly.
By comparing our method with these two baselines, we can prove that learning common features and modality specific features can improve the segmentation accuracy.
In the third baseline (C-DN), we train a deconvolutional network that takes the four-channel RGB+depth as the input.
By comparing our method with this baseline, we can show that explicitly disentangling the common and specific features can improve the segmentation accuracy.
In the fourth baseline (E-DN), we use the proposed network structure and Euclidean distance to assess the similarity between common (or modality specific) features individually.
By comparing our method with this baseline, we can prove that MK-MMD is better than Euclidean distance, as a measurement to assess the similarity between feature distributions.
The fifth baseline (U-DN) is the unregularized version of our framework. In this baseline, the loss function only has the first two terms of Eq. \ref{eq:lossfunction}, and the last two terms are removed.

\begin{figure}
	\begin{center}
		\includegraphics[width=0.8\linewidth]{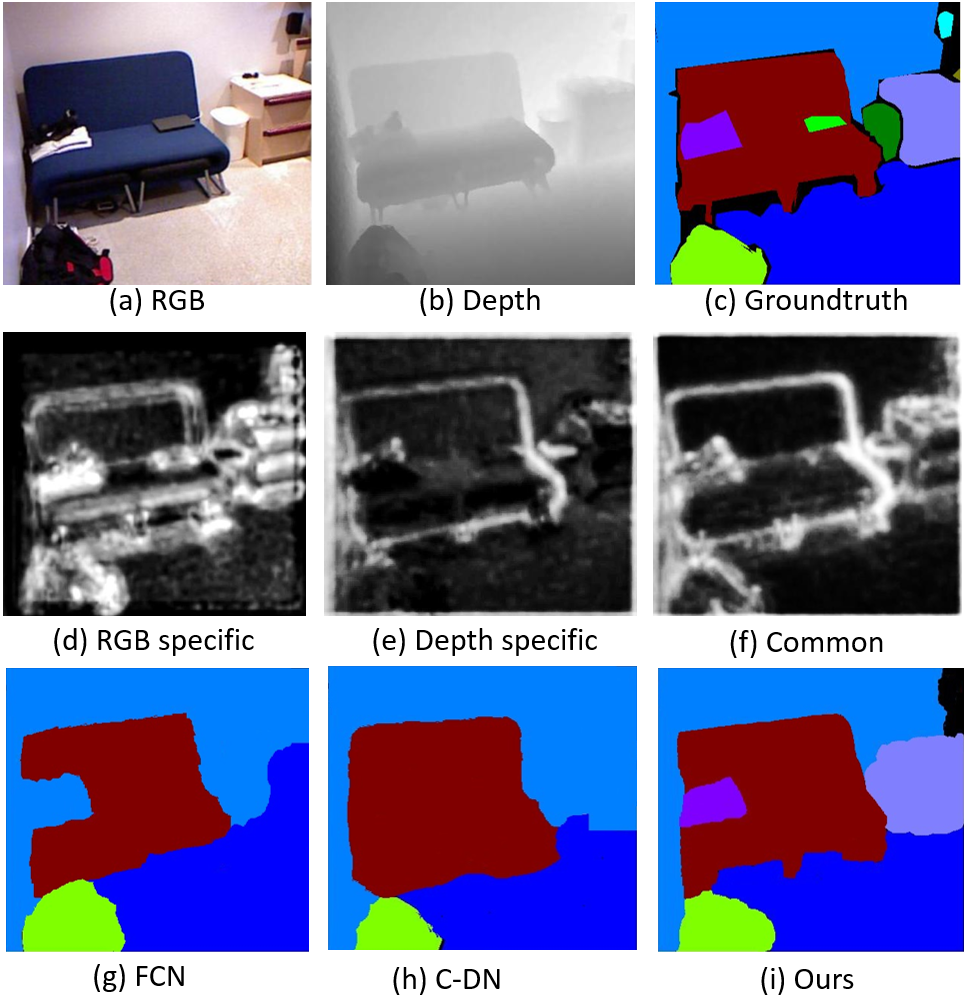}
	\end{center}
	\caption{An example image from NYU depth dataset V2 \cite{Silberman:ECCV2012:Indoor_Segmentation}. The The first row shows (a) RGB image, (b) depth map, and the (c) ground truth. The second row shows the learned (d) RGB specific features, (e) depth specific features, and (f) common features. The third rows shows the segmentation results of two based lines ((g) FCN and (h) C-DN) and (i) our method}
	\label{fig:expermental-results}
\end{figure}

\subsection{Testing}
\label{sec:experiment-testing}

For a testing image, we first generate $ 100 $ patches or bounding boxes \cite{ZitnickDollar_ECCV14_edgeBoxes}, each of them corresponding to a potential object. 
Then, we segment these patches individually using the learned network. Finally, we combine the segmentation results of these patches by decision score fusion to obtain the final label map.

We first visualize the segmentation results, as well as the learned common and specific features in Fig \ref{fig:expermental-results}. In this figure, we also show the segmentation results of FCN and the baseline C-DN (that takes the concatenated RGB-D as input). Both of these comparison methods can not segment the table correctly.
However, using in our method, the RGB specific and common features can characterize the table correctly.
The second row of Fig \ref{fig:expermental-results} show the feature maps of deconvolutional layers. For visualization of RGB specific features, we only take the $ s_{rgb} $ as the input of layer $ fc2_{rgb} $ and ignore the common features. The depth specific features are visualized in the similar way. To show the common feature, we drop the specific feature and only take the common features $ c_{rgb} $ and $ c_d $ as the input of the layer $ fc2_{rgb} $. The features in Fig. \ref{fig:expermental-results} are the feature maps of layer deconv 2-2.  While the RGB specific features mainly characterize the texture-rich regions, the depth specific features characterize the edges of objects.

\begin{table}
	\caption{The average 13-class segmentation accuracy of different methods on the NYU depth dataset V1 (KDES represents kernel descriptor) }
	\begin{center}
		\begin{tabular}{ | l | c | c | l| c|}
			\hline Method & Acc & & Method & Acc \\
			\hline    
			Silberman and Fergus \cite{Silberman_ICCV2011_Indoor_scene} & 53.0\% & &  Pei et al. \cite{Pei_IJcnn-unsupervisedMultimodal} & 50.5\%
			\\ \hline
			
			Wang et al. \cite{Wang_ECCV2014_multi-modal-unsupervised} & 72.9\%  &  &
			Hermans \cite{Hermans14ICRA} & 59.5\%
			\\ \hline
			KDES-RGB \cite{Ren_Xiaofeng_CVPR2012_RGB-D_SceneLabeling} & $66.2\% $ & &
			KDES-depth \cite{Ren_Xiaofeng_CVPR2012_RGB-D_SceneLabeling} & $63.4\% $ \\
			\hline
			KDES RGB-D \cite{Ren_Xiaofeng_CVPR2012_RGB-D_SceneLabeling} & $71.4\% $  & &
			KDES Treepath \cite{Ren_Xiaofeng_CVPR2012_RGB-D_SceneLabeling} & $74.6\% $ \\
			\hline
			KDES MRF \cite{Ren_Xiaofeng_CVPR2012_RGB-D_SceneLabeling} & $74.6\% $ & &
			KDES Tree+MRF \cite{Ren_Xiaofeng_CVPR2012_RGB-D_SceneLabeling} & $76.1\% $ \\ \hline

			B-DN & 76.5\% & & S-DN & 72.1\% 
			\\ \hline
			C-DN & 70.3\% & & E-DN & 71.4\% 
			\\ \hline
			U-DN & 69.9\% & & Ours & 78.8\% 
			\\ \hline
			
		\end{tabular}
		\label{tab:NYU-v1-accuracy}
	\end{center}
\end{table}

Tab. \ref{tab:NYU-v1-accuracy} lists the class average accuracies of different methods on the NYU depth dataset V1. We compare our method with five baselines and five different works \cite{Ren_Xiaofeng_CVPR2012_RGB-D_SceneLabeling},  \cite{Silberman_ICCV2011_Indoor_scene},  \cite{Pei_IJcnn-unsupervisedMultimodal},  \cite{Wang_ECCV2014_multi-modal-unsupervised}, \cite{Hermans14ICRA}. The methods  proposed in \cite{Ren_Xiaofeng_CVPR2012_RGB-D_SceneLabeling} and \cite{Silberman_ICCV2011_Indoor_scene} use hand-craft features. Pei et al. \cite{Pei_IJcnn-unsupervisedMultimodal} use a one-layer network and Wang et al. \cite{Wang_ECCV2014_multi-modal-unsupervised} use a two-layer network to learn representations. Our deep network can achieve higher accuracy than their methods. It indicates that deep features can perform better than shallow features.

Tab. \ref{tab:NYU-v2-accuracy} lists the accuracies of the proposed method and four baselines, as well as the methods proposed in \cite{couprie_2013ICLR_indoor_semantic} and \cite{Wang_ECCV2014_multi-modal-unsupervised} on the dataset NYU V2. 
We can see from Tab. \ref{tab:NYU-v2-accuracy} that the proposed method outperforms the previous state-of-the-art \cite{Wang_ECCV2014_multi-modal-unsupervised} by $ 6.3\% $. Notably, in the classes of \textit{objects}, \textit{furniture} and \textit{decorate}, our method significantly outperforms  \cite{Wang_ECCV2014_multi-modal-unsupervised}. In Tab. \ref{tab:4-13-40-NYU-v2-accuracy}, we compare our method with the previous works on the 4-class, 13-class, and 40-class segmentation task. The proposed method outperforms all of them in segmentation accuracy.

\begin{table}
	\caption{The 13-class segmentation accuracy of different methods on NYU depth dataset V2. B-DN trains two independent deconvolutional networks. S-DN contains a single deconvolutional network and two convolutional networks. C-DN trains one convolutional and one deconvolutional network with 4-channel RGBD as input. E-DN uses the Euclidean distance to assess the difference between features in the proposed network}
	\begin{center}
		\begin{tabular}{ | l | c | c | c | c | c |c |c|}   
			
			\hline    
			&   Couprie \cite{couprie_2013ICLR_indoor_semantic}  & Wang \cite{Wang_ECCV2014_multi-modal-unsupervised}  &  B-DN &  S-DN & C-DN& E-DN &
			Ours \\ 
			\hline
			bed & 38.1 &  47.6 & 27.4 & 19.6 & 19.2 & 25.3 & 31.6 \\ \hline
			objects & 8.7 &  12.4 &  40.7& 43.8& 40.1& 36.9 & 61.5 \\ \hline
			chair & 34.1 &  23.5 & 43.5& 39.2& 42.8& 39.3& 43.6 \\ \hline
			furniture & 42.4 &  16.7 & 37.2& 36.3 & 35.7& 35.2& 49.8 \\ \hline
			ceiling & 62.6 &  68.1 & 52.2 & 56.2 & 55.9 & 55.1& 58.7 \\ \hline
			floor & 87.3 &  84.1 & 82.9& 86.5& 84.7 & 90.5& 89.0 \\ \hline
			decorate & 40.4 &  26.4 & 55.8 & 56.9 & 54.6 & 60.4& 68.9 \\ \hline
			sofa & 24.6 &  39.1 & 36.7 & 31.3 & 28.3& 35.7& 30.8 \\ \hline
			table & 10.2 &  35.4 & 36.4& 50.3& 50.5& 42.7& 49.3 \\ \hline
			wall & 86.1 & 65.9 & 41.4& 32.2& 33.3 & 34.3& 44.9 \\ \hline
			{\small  window} & 15.9 &  52.2 & 81.7& 87.4& 88.9& 78.1& 83.9 \\ \hline
			books & 13.7 &  45.0 & 28.8& 23.1& 22.5& 29.9& 39.9 \\ \hline 
			TV & 6.0 &  32.4 & 53.7& 29.4& 29.7 & 35.27& 32.8 \\ \hline \hline
			\textit{AVE} & 36.2 &  42.2 & 47.6&  45.6&  45.1&  46.1&  52.7 \\ \hline
		\end{tabular}
		\label{tab:NYU-v2-accuracy}
	\end{center}
\end{table}

\begin{table}
	\caption{The per-class accuracy of 4, 13, and 40-class segmentation on NYU depth dataset V2}
	\begin{center}
		\begin{tabular}{ | c | c |c | c | c |c| c | c |}   
			
			\hline 
			\multicolumn{2}{|c|}{4-class} & & \multicolumn{2}{|c|}{13-class} & & \multicolumn{2}{|c|}{40-class} \\ \hline \hline
			Couprie \cite{couprie_2013ICLR_indoor_semantic} & 63.5\% & & Couprie \cite{couprie_2013ICLR_indoor_semantic}  &  36.2\% & & Gupta'13 \cite{Gupta_CVPR2013_Perceptual_Organization} & 28.4\%\\ \hline
			
			Khan \cite{Khan_ECCV2014_Geometry} &  65.6\% &  & Wang \cite{Wang_ECCV2014_multi-modal-unsupervised} & 42.2\% & & Gupta'14 \cite{gupta_ECCV14_Learning_rich} & 35.1\%  \\
			\hline
			
			Stuckler \cite{Stuckler:2015:DRM:2849459.2849485} & 67.0\% & & Khan \cite{Khan_ECCV2014_Geometry} & 45.1\% & & Long \cite{long_shelhamer_CVPR2015_fcn} & 46.1\% 
			\\
			\hline
			Muller \cite{muller2014learning} & 71.9\% & & Hermans \cite{Hermans14ICRA} & 48.0\% & & Eigen \cite{eigen2015predicting} & 45.1\% \\
			\hline
			
			U-DN & 71.8\% & & U-DN & 49.2\% && U-DN & 41.7\% \\
			
			\hline \hline
			Ours &  74.7\% & & Ours & 52.7\% & & Ours & 47.3\% \\ \hline
		\end{tabular}
		\label{tab:4-13-40-NYU-v2-accuracy}
	\end{center}
\end{table}

Based on Tab. \ref{tab:NYU-v1-accuracy} and Tab. \ref{tab:NYU-v2-accuracy}, we can conclude that MK-MMD is a better measurement than Euclidean distance to learn common features and modality specific features in this task. The proposed method outperforms the baseline E-DN by 7.4\% and 6.6\% in dataset V1 and V2, respectively. This is mainly because, the Euclidean distance is heavily effected by the outliers.

In both B-DN and the proposed network, we use a linear combination to conduct decision score fusion. Compared with the baseline B-DN, the proposed network is 2.3\% higher on dataset V1 and 5.1\% higher on dataset V2.
It indicates that we should correlate the two modalities in the feature learning stage instead of only fusing them at the decision score level. 
The baseline B-DN is not robust. Its segmentation accuracy varies a lot as the parameter (for linear combination of decision score fusion) changes. By contrast, our network is much more robust and varies slightly as the parameter changes. In our network, the deconvolutional results from two modalities are much more similar than those in B-DN. 
The reason is that, by borrowing common features from other modality, our method will produce similar decisions scores for the two modalities, which makes fusing result robust to the linear combination parameter.

\section{Conclusion}
\label{sec:conclusion}

In this paper, we propose a new network structure for RGB-D semantic segmentation. The proposed network has a convolutional network and a deconvolutional network for each of the modality. We bridge the convolutional networks and the deconvolutional networks using a feature transformation network. In the feature transformation network, we transform the convolutional features into common features and modality-specific features. Instead of using a one-vs-one strategy to measure the similarity between features, we adopt MK-MMD to calculate the similarity between their distributions. To learn robust deconvolutional features, we allow one modality to borrow the common features from the other modality. Our method achieves competitive performance on NYU depth dataset V1 and V2.

\section*{Acknowledgment}

The research is supported by Singapore Ministry of Education (MOE) Tier 2 ARC28/14, and Singapore A*STAR Science and Engineering Research Council PSF1321202099. The research is also supported by Australian Research Council Projects DP-140102164, FT-130101457, and LE-140100061.

This work was carried out at the Rapid-Rich Object Search
(ROSE) Lab at the Nanyang Technological University, Singapore.
The ROSE Lab is supported by a grant from the
Singapore National Research Foundation and administered by
the Interactive \& Digital Media Programme Office at the
Media Development Authority.

\bibliographystyle{splncs}
\bibliography{camera_ready_bib}
\end{document}